%% file: article.tex
\documentclass[journal]{IEEEtran}

\usepackage{graphicx}
\graphicspath{ {images/} }

\usepackage[colorinlistoftodos,prependcaption,textsize=tiny,textwidth=1.3cm]{todonotes}
\presetkeys{todonotes}{backgroundcolor={yellow!50},fancyline}{}

\usepackage{todonotes}

\usepackage{times}

\usepackage{amssymb}
\usepackage{amsmath} 

\usepackage{bm}
\usepackage{bbm}

\usepackage{textgreek}

\usepackage[inline]{enumitem}

\DeclareGraphicsExtensions{.png,.pdf,.jpeg}

\ifCLASSOPTIONcompsoc
  \usepackage[nocompress]{cite}
\else
  \usepackage{cite}
\fi
\ifCLASSINFOpdf
\else
\fi
\ifCLASSOPTIONcompsoc
 \usepackage[caption=false,font=footnotesize,labelfont=sf,textfont=sf]{subfig}
\else
 \usepackage[caption=false,font=footnotesize]{subfig}
\fi
\newcommand{\mathbm}[1]{\bm{#1}}

\begin{document}
%
\title{mGNN: Generalizing the Graph Neural Networks to the Multilayer Case} 
%
%
%
%

\author{Marco Grassia,
        Manlio De Domenico,
        Giuseppe Mangioni
\IEEEcompsocitemizethanks{%
\IEEEcompsocthanksitem M. Grassia and G. Mangioni are with the Department
of Electric Electronic and Information Engineering, University of Catania, Italy.\\
E-mail: marco.grassia@unict.it, giuseppe.mangioni@unict.it.
\IEEEcompsocthanksitem  M. De Domenico is with CoMuNe Lab, Fondazione Bruno Kessler, Via Sommarive 18, 38123 Povo (TN), Italy.\\
E-mail: mdedomenico@fbk.eu.
}
%
}

\IEEEpubid{\makebox[\columnwidth]{\hfill 0000--0000/00/\$00.00~\copyright~2021 IEEE}%
\hspace{\columnsep}\makebox[\columnwidth]{Submitted to the IEEE Computer Society\hfill}}


\IEEEtitleabstractindextext{%
\begin{abstract}
Networks are a powerful tool to model complex systems, and the definition of many Graph Neural Networks (GNN), Deep Learning algorithms that can handle networks, has opened a new way to approach many real-world problems that would be hardly or even untractable.
In this paper, we propose \emph{mGNN}, a framework meant to generalize GNNs to the case of multi-layer networks, i.e., networks that can model multiple kinds of interactions and relations between nodes.
Our approach is general (i.e., not task specific) and has the advantage of extending any type of GNN without any computational overhead.
We test the framework into three different tasks (node and network classification, link prediction) to validate it.
\end{abstract}

}

\maketitle


\IEEEdisplaynontitleabstractindextext

%
\IEEEpeerreviewmaketitle

\input{introduction}

\input{related}

\section{Preliminaries}
\input{gnns}
\input{multilayer}

\input{framework}
\input{results}

\input{conclusions}
\appendices

\section{Training and model parameters}

\subsection{Malaria genes classification}
The model we use to classify the nodes has $6$ supra-layers with identical sub-layers (i.e., \emph{GAT} layers with $60$ output features and $5$ heads each, negative slope $0.2$), followed by a Multi-Layer Perceptron with $6$ outputs (the classifier) that takes as input the features of the replicas of each node and predicts its class.
The model is trained for $2500$ epochs (with $100$ epochs patience) with learning rate $5 \cdot 10^{-4}$, weight decay $10^{-3}$ and $0.3$ dropout probability.
We use the Cross Entropy loss function with a rescaling weight to account for the different distribution of classes.

\subsection{Link prediction}
The model we use to perform link prediction has $3$ identical supra-layers (i.e., \emph{GAT} sub-layers with $30$ output features and $5$ heads each, negative slope $0.2$).
We train the model for $2500$ epochs (with $400$ epochs patience) with learning rate $10^{-3}$, weight decay $10^{-5}$ and $0.3$ dropout probability.

\subsection{Superdiffusion prediction}
The model we use for this task includes $4$ supra-layers with identical sub-layers (i.e., $GAT$ layers with $10$ output features and $5$ heads each, negative slope $0.2$) and is trained using the Mean Square Error loss function for $100$ epochs with learning rate $5 \cdot 10^{-3}$, weight decay $10^{-5}$ and $0.3$ dropout probability.
The node embeddings are aggregated using a weighted sum function learned during the training.



\ifCLASSOPTIONcaptionsoff
  \newpage
\fi



%



\bibliographystyle{IEEEtran}
\bibliography{article}

%

\begin{IEEEbiography}[{\includegraphics[width=1in,height=1.25in,clip,keepaspectratio]{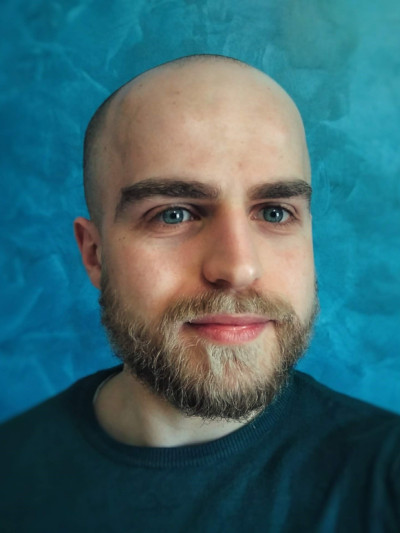}}]{Marco Grassia} is a Ph.D. student in Network Science and Machine Learning at Department of
Electric, Electronic Informatics, University of Catania, where he graduated with a MS in Computer Engineering in 2018.
His research interests include Network Science and the application of Geometric Deep Learning techniques to approach computationally challenging problems on networks.
\end{IEEEbiography}

\begin{IEEEbiography}[{\includegraphics[width=1in,height=1.25in,clip,keepaspectratio]{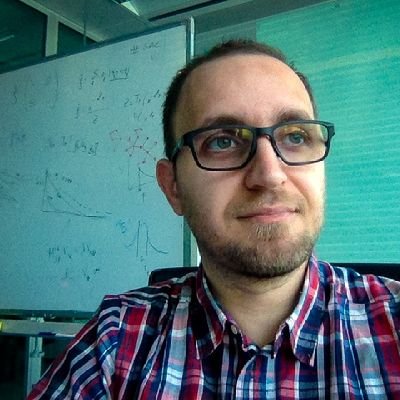}}]{Manlio De Domenico.}
Senior Researcher and Head of the Complex Multilayer Networks (CoMuNe) Research Unit at the Center for Digital Society of Fondazione Bruno Kessler. His research activity is at the edge of theoretical, experimental and computational aspects of statistical physics of complex systems, with focus on multilayer network modeling and emergent phenomena in biological, ecological, socio-technical and socio-ecological ones.
\end{IEEEbiography}

\begin{IEEEbiography}[{\includegraphics[width=1in,height=1.25in,clip,keepaspectratio]{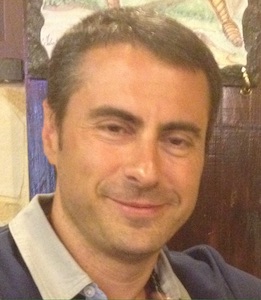}}]{Giuseppe Mangioni.} 
Giuseppe Mangioni is associate professor in Department of
Electrical Electronics and Computer Engineering at the University of Catania.
He received the degree in Computer Engineering (1995) and
the Ph.D. degree (2000) at the University of Catania. Currently
he teaches advanced programming languages, internet architecture and OOP at the University of Catania. His research interests include network science, trust and reputation systems, machine learning, peer-to-peer systems,  self-organizing and
self-adaptive systems. He is co-initiator and member of the steering committee of the International Conference on Complex Networks (CompleNet).
\end{IEEEbiography}







\end{document}

%% file: introduction.tex
\section{Introduction}\label{sec:introduction}

Complex networks, the abstract representation of interactions and relationships between the constituents complex systems, are nowadays pervasive in many fields of science. The versatility of networks for mapping the complexity of a broad variety of empirical systems, makes them suitable mathematical objects granting a wide applicability range.
The success of network modeling has been widely demonstrated with applications in biology\cite{jeong2000large}, ecology\cite{williams2002two}, neuroscience\cite{varela2001brainweb}, economy\cite{schweitzer2009economic}, finance\cite{allen2009networks}, engineering\cite{cardillo2006structural}, physics\cite{albert2002statistical}, social\cite{vega2007complex} and computer science\cite{faloutsos2011power}, where units can be proteins, animal species, neurons, people or machines, and links can encode regulatory interactions, predator-prey interactions, synaptic connections, social relationships or communication channels, just to mention a few emblematic examples.
This broad applicability to problems of empirical interest has motivated scientist to invest considerable research efforts in developing new theories and tools, making networks more and more powerful, in terms of both descriptive and prescriptive power, as well as scalability.

In fact, many problems on networks require a high computational effort to be solved, and, often, such problems need approximated algorithms to be tractable in real-world situations.
This issue has stimulated scientists to develop new techniques inspired to those employed in the machine learning field, where complex problems have been successfully solved by using deep--learning models and algorithms.
Recently, \emph{geometric deep learning}~\cite{bronstein2017geometric}, a completely new branch of research aimed at enabling the application of deep--learning to networks (and to non--euclidean spaces in general), has been developed, offering up new research opportunities and tools to tackle complex problems, as shown in~\cite{hamilton2017representation}. 
This has led, in recent years, to the development of many Graph Neural Networks (GNNs)~\cite{4700287} models (see section~\ref{s:gnns} for details) that differ mainly in the type of problems they are meant to solve and/or in their performance, as described in the section~\ref{s:related_works}.

However, it has been shown in the last decade that while networks are wide applicable, many empirical systems exhibit multiple types of interactions or relationships simultaneously, introducing a new level of complexity and topological correlations~\cite{nicosia2015measuring} which required the development of an \emph{ad hoc} mathematical framework~\cite{de2013mathematical} to deal with. For instance, in the case of transportation systems places might be connected by different transportation means such as train, bus, tube, etc~\cite{de2014navigability,aleta2017multilayer}. Another emblematic example concerns social systems, where people can exhibit several different kind of relations, such as friendship, acquaintance or business, etc~\cite{dickison2016multilayer}.
Furthermore, international trade systems can also be characterized by the presence of several kind of relationships among countries, depending on the commodities they trade~\cite{bhattacharya2008international,barigozzi2011identifying}.
Also, many biological systems are well represented by multiple types of relationship among their constituents, such, for instance, in the Homo Sapiens proteome, where protein--protein relations can be of two types, physical and genetic, such as interactions, chemical associations or post-translational modifications~\cite{mangioni2018multilayer,choobdar2019assessment,verstraete2020covmulnet19}.
We refer the reader to~\cite{gao2012networks,boccaletti2014structure,kivela2014multilayer,wang2015evolutionary,de2016physics} for more thorough reviews about multilayer and interdependent systems and other examples of applications.

A simple way to study systems characterized by multiple relationships is to consider a single network (or monoplex) that is the result of the aggregation of the different kinds of relations.
While this approach has been used often in the past, it has multiple critical issues and received a lot of criticism, since it is inherently affected by loss of potentially critical information about the structure of the system and, consequently, about its function.
For example, some of the open questions are: \emph{how to aggregate the layers?}, and, \emph{is it correct to do so?}.
The answer to the first question requires the definition of an aggregation function (sum, average, min--max, etc.) and a proper scaling of the weights associated to each kind of relation.
\IEEEpubidadjcol
The second question poses an even more important problem: does it conceptually make sense to aggregate?
The answer, in general, is negative, since often relations encode different contexts that cannot be simply mixed together without altering the structure or the function of the system under investigation~\cite{sola2013eigenvector,cardillo2013modeling,de2014navigability,battiston2014structural,estrada2014communicability,de2015ranking,de2015structural,ghavasieh2020enhancing}.


Similarly to the monoplex case, many problems on multi-layer networks are computationally hard to solve.
But, unlike in the case of monoplex networks, there are no geometric deep--learning tools suitable for multi-layer networks, a significant knowledge gap in the literature.
In this paper we propose \emph{mGNN} an extension of Graph Neural Networks to deal with multi-layer networks.
Specifically, we propose a framework able to manage both the intra- and inter--layer relations of these networks by using any kind of GNN layers.
To validate our framework, we show its application to solve three real--world problems on multi-layer networks: nodes (genes) classification in a genetic multi-layer network related to malaria, link prediction in a multiplex social network (FriendFeed, Twitter and YouTube users) and multi-layer network classification in a super-diffusion~\cite{gomez2013diffusion,sole2013spectral} prediction problem.

The paper is organized as in the following.
We introduce the fundamentals of Graph Neural Networks and some of their applications in Section~\ref{s:gnns}, and the fundamentals of multi-layer networks in Section~\ref{s:multilayers}. 
In Section~\ref{s:related_works} we discuss the related works.
Section~\ref{s:framework} describes our framework and Section~\ref{s:results} presents the results.
Finally, Section~\ref{sec:conclusions} concludes the paper.

%% file: related.tex
\section{Related Works}
\label{s:related_works}



The first proposal of Machine Learning techniques for graph-structured data dates back to 2009 with the work of Scarselli et al.~\cite{4700287}, but only in the last few years, after the explosion in popularity of Deep Learning, the interest in such techniques grew. 
In fact, the lack of Deep Learning algorithms for non-Euclidean data led the efforts of many researchers and multiple attempts to bridge Deep Learning and graphs have been made, most of which share the key intuition of generalizing the convolution operator to graphs.
The rationale of this choice is that the convolution operator --- at the core of \emph{Convolutional Neural Networks} (CNNs) --- is a local operator with nice properties (such as low computational complexity, shared parameters, etc.) that performs incredibly well in many fields like image recognition.
One of the first works in this direction is \emph{Graph Convolutional Networks} (GCNs) by Kipf et al.~\cite{kipf2016semi}, but many others followed, like \emph{GraphSAGE} by Hamilton et al. and like \emph{Graph Attention Networks} (GAT)~\cite{gatlayers} by Veličković et al. that borrow the \emph{attention mechanism} from the Natural Language Processing (NLP) field to assign importance weights to each neighbouring node.
These algorithms are usually called Graph Neural Networks (GNNs), in analogy to their Euclidean ancestor.

If there are remarkable works in the literature about the definition of new GNN algorithms, even more works can be found with disparate applications.
For instance, GNNs have been used:
\begin{itemize}
    \item for drug discovery and development by Gaudelet et al.~\cite{gaudelet2021utilising};
    \item to predict poly-pharmacy side effects by Zitnik et al.~\cite{Zitnik2018};
    \item to build recommender systems, like \emph{PinSage}, deployed at Pinterest, by Ying et al.~\cite{Ying_2018};
    \item to approach the Network Dismantling problem and to provide an Early Warning signal of system disintegration by Grassia et al.~\cite{grassia2021machine};
    \item for chip placement by Mirhoseini et al.~\cite{mirhoseini2020chip}.
\end{itemize}


While GNN layers are meant to work with graphs, they cannot be used to process multi-layer networks directly and, considering their importance and ubiquity, further work is needed to overcome this limitation.
In fact, in the literature there are two prevalent ways to approach multi-layer networks with GNNs: 
\begin{enumerate*}
    \item An aggregate-all layers approach (i.e., compressing the multi-layer network in a single graph that can be feed to a GNN). This leads to the loss of useful information of the inter-layer connections and of the different meaning/dynamics of each layer;
    \item Tailoring the approach to the specific problem addressed, with the main limitation being the need of defining new methodology that cannot generalize well to other problems and that does not answer the general question of how to use GNNs in the setting of multi-layer networks.
\end{enumerate*}

However, some attempts in generalizing GNNs to the multi-layer case have been made, and here we analyze the most promising ones and their limitations:
\begin{itemize}
    \item \emph{Multi-Layered Network Embedding} (MANE) by Li et al.~\cite{MANE}.
        In their work, in order to get the multi-layer node embeddings, they optimize an objective function that is the difference of two terms: an intra-layer term, where they optimize the embeddings of nodes in each layer so that neighbouring nodes are close to each other (that is, no input node feature is supported) and an inter-layer term, meant to make the embeddings of nodes in different layers close to each other if these layers depend on each other, according to the user-defined dependencies.
        %
        The main limitations of this approach are:
        \begin{enumerate*}
            \item trained models do not generalize to unseen networks;
            \item it does not support input node features and the output embeddings are topology-based only;
            \item the need of a custom weight to balance the intra- and inter- layer contribution to the objective function;
            \item the need of defining layer-layer dependencies;
            \item the high computational complexity ($\#iters \cdot \mathcal{O}(N^2)$, where $\#iters$ is the number of iterations needed for convergence).
        \end{enumerate*}
    \item \emph{Semi-supervised Classification in Multi-layer Graphs with Graph Convolutional Networks} (MGCN): in this work, Ghorbani et al.~\cite{Ghorbani_2019}, propose a semi-supervised multi-layer node embedding framework.
        Specifically, they employ \emph{Graph Convolutional Networks} (GCN) by Kipf et al. to process layers individually (i.e., a GCN per layer that does not account for inter-layer connections) and get the node embeddings, which are trained by optimizing a custom loss function composed of two parts: an unsupervised part, used to train the GCNs by reconstructing both the intra- and the inter-layer connections, and a supervised part that trains the GCNs on node classification task using node-labels for a sub-set of nodes.
        The authors compare against MANE, showing better performance.
        The most important limitations of this approach are:
        \begin{enumerate*}
            \item the need of training the GCNs to reconstruct the network, which is not useful in many tasks and creates an overhead;
            \item the node embeddings are computed on each layer independently, thus the feature propagation happens only in the intra-layer;
            \item it needs a custom loss function for each task and to choose a custom weight to sum the two terms of the loss.
        \end{enumerate*}
        %
    %
    \item \emph{Multi-GCN: Graph Convolutional Networks for Multi-View Networks, with Applications to Global Poverty} by Raza Khan et al.~\cite{khan2019multigcn}, who focus on multiplex (multi-view) networks by proposing a method to "fuse the multiple views of a graph".
        In particular, their approach consists of three steps:
        they first employ sub-space analysis to merge the layers of the network, then they identify the most informative sub-components via a manifold ranking procedure and, finally, they feed the resulting matrix to a Convolutional Neural Network (CNN) adapted to graph-structured data. 
        The main limitations are:
        \begin{enumerate*}
            \item the focus on multiplex networks;
            \item the large overhead to fuse the layers and feed the result to an adapted CNN;
            \item the extremely high computational complexity ($\mathcal{O}(N^3)$);
            \item the small performance margin over mono-plex GNNs.
        \end{enumerate*}
\end{itemize}


%% file: gnns.tex
\subsection{Graph Neural Networks}
\label{s:gnns}
In this section, we introduce the reader to the basics of the Graph Neural Networks (GNNs) and the notation used in this paper.

The idea behind Graph Neural Networks is to generalize to graph-structured data the convolution operator at the core of the Convolutional Neural Networks (CNNs) and the success of Deep Learning.
In fact, while the CNNs are well defined for Euclidean data (images are grids of pixels), the convolution needs a new formulation for non-Euclidean domains like graphs and manifolds.

In analogy with the Euclidean case where, in each convolutional layer, a kernel (i.e. a learned tensor) is used to multiply and aggregate the neighbourhood of each pixel, the graph convolution operator should aggregate the features of the neighbours of each node using some learned function.
In both cases, the output values provided by the aggregation are higher level features that will be fed to the next layer of the model. 
Thus, by stacking many layers together, one should get higher and higher level features that capture some high-order pattern.
In the field of Geometric Deep Learning, these high-level features are also called \emph{node embedding}, i.e., a node is mapped into an Euclidean space where classical Machine Learning tools, like Multilayer Perceptrons, can be used to perform classification, regression, etc.

In the last years, many GNN layers have been proposed by researchers and the interest in the field keeps growing thanks to the large availability of graph structured data and to the many interesting applications.
For instance, GNNs have been used to perform node classification or regression --- e.g., classifying entities like users or assigning them a score ---, but also to perform graph classification or regression (by aggregating all the node embeddings together) and even to perform link prediction --- i.e., the prediction of whether a link (modeling a relation, interaction, etc.) exists between two nodes (e.g., users, molecules, etc.).
Another reason behind the success of GNNs is that, just like their Euclidean counterpart, they usually have low computational complexity, thanks to the locality of the convolutional operator, and are able to generalize to previously unseen nodes and graphs (\emph{inductive} capability).

\begin{figure*}[!ht]
    \centering
    \includegraphics[width=0.8\textwidth]{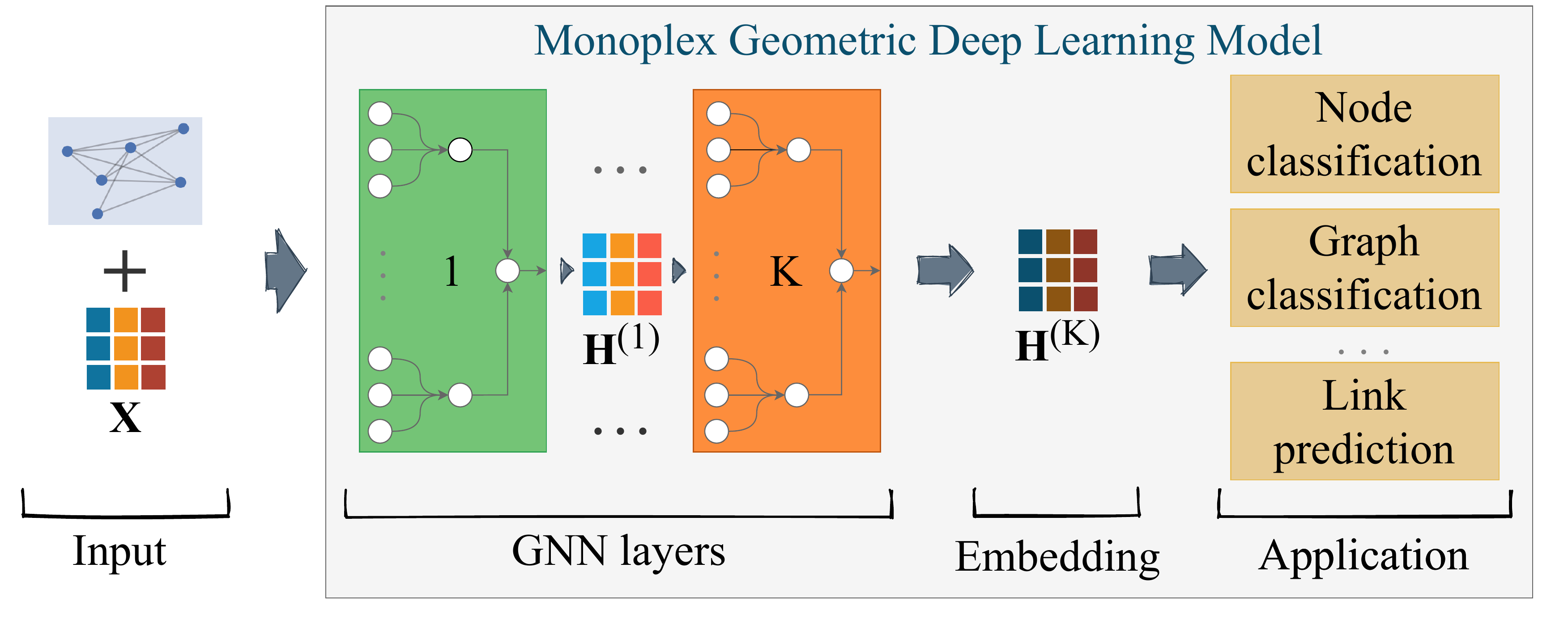}
    \caption{\textbf{A Geometric Deep Learning model example.} $K$ is the number of convolutional layers, $\mathbf{X}$ are the input node features. The node embeddings $\mathbf{H}^{(K)}$, can be used in various applications.}
    \label{f:gcn_monoplex}
\end{figure*}

In this paper, we employ the following notation, common in the literature, and indicate the features of node $i$ after $k$ graph convolutional layers with $\mathbf{h}_{i}^{(k)}$.
According to this notation, $\mathbf{h}_{i}^{(0)}$ are the input node features of node $i$ ($\mathbf{x}_{i}$), and being $K$ the number of convolutional layers in the model, $\mathbf{h}_{i}^{(K)}$ will be its output features ($i$'s embedding).
We show the basic structure of a Geometric Deep Learning model in \figurename~\ref{f:gcn_monoplex}.
The model takes as input the network and its node features, which are fed to the first convolutional layer that produces $\mathbf{H}^{(1)}$, the tensor of higher-level features of all the nodes, that are fed, in turn, to the next convolutional layer.
This process repeats for all $K$ layers and the output features $\mathbf{H}^{(K)}$ are the (output) node embedding that can be used for the user's application.
More formally, a (basic) generic GNN layer can be defined as:
\begin{equation*}
    \mathbf{h}_{i}^{(k+1)} = AGG_{k+1}(\mathbf{h}_{i}^{(k)}, \{\mathbf{h}_{j}^{(k)}, \forall j \in \mathcal{N}^{in}_i\})
\end{equation*}

where $AGG_{k+1}$ is a generic learned aggregation function and $\mathcal{N}_i$ is the neighbourhood of node $i$.

From a network perspective, the computation of node $i$'s embedding $\mathbf{h}_{i}^{(K)}$ is the propagation of the node features in a bottom-up fashion in a tree where the root is the node $i$ itself and the nodes at each layer $k$ are $i$'s $k$-hops neighbours.

For more about Graph Neural Networks, we refer the reader to~\cite{grl-book}.




To avoid confusion between neural networks and networks (graphs), in this paper we will refer to the first as "neural networks" and to the latter as networks/graphs, and to avoid confusion between model layers and network layers, we will refer to each by specifying the context.







%% file: multilayer.tex
\subsection{Multilayer Networks}
\label{s:multilayers}

In the following, we introduce the fundamental concepts defining multilayer networks.

Classical networks are usually represented by their adjacency matrix $\mathbf{W}$ whose entries $W_{ij}$ encode the presence (or the absence) of a link between nodes $i$ and $j$ ($i,j=1,2,...,N$, where $N$ is the systems' size)~\cite{newman2010networks}. However, in empirical systems, two nodes can often interact in multiple distinct ways: e.g., two individuals might have distinct type of social relationships (friendship, trust, business, family, so forth and so on), two areas of a city might be connected by distinct types of transportation means (rail, tube, bus, etc), two proteins within a cell might have physical or functional interactions (inhibitory, excitatory, etc).

Unfortunately, the classical representation is no more enough to account for such an additional level of complexity and a new mathematical framework is needed. Let us assume that the same pair of nodes $i$ and $j$ can be related in $L$ distinct ways at most, i.e., we can encode their multiple relationships with a vector $(W_{ij}^{(1)},W_{ij}^{(2)},...,W_{ij}^{(L)})$ of non-negative numbers. In the following, each way encodes a network layer.

Note that this representation is not enough, since one might want to encode interactions between a node $i$ in a layer $\alpha$ and a node $j$ in a layer $\beta$. To this aim, it can be shown that one can reliably account for this complexity by introducing a higher-order matrix, more specifically a rank-4 adjacency tensor, with components $M^{i\alpha}_{j\beta}$, where $\alpha,\beta=1,2,...,L$~\cite{de2013mathematical}. It is straightforward to show that for $\beta=\alpha$ one retrieves all interactions between nodes only in layer $\alpha$, thus defining intra-layer connectivity. Conversely, inter-layer connectivity is obtained for $\beta\neq\alpha$. It is possible to exploit tensorial algebra to generalize many classical network descriptors, from centrality measures to community detection (see \cite{mucha2010community,de2013mathematical,kivela2014multilayer,boccaletti2014structure,de2015ranking,bazzi2016community} and references therein). When inter-layer connectivity can not be defined from data, it is still possible to analyze this type of networks as edge-colored multigraphs which, in turn, require a rank-3 tensor for their mathematical representation~\cite{bianconi2013statistical,battiston2014structural}.

One of the applications of this work deals with diffusion, a physical process whose dynamics on classical graphs has been extensively studied~\cite{masuda2017random} and known to be governed by a master equation where the combinatorial Laplacian matrix (or other types of Laplacian matrices) allows to define the underlying propagator. The generalization of master equations to describe diffusion processes in multilayer networks allowed to define the corresponding multilayer Laplacian tensors~\cite{gomez2013diffusion,de2013mathematical,de2014navigability,de2016physics} $\mathcal{L}^{i\alpha}_{j\beta}$ which, in the case of continuous diffusion takes the form
\begin{eqnarray}
\mathcal{L}^{i\alpha}_{j\beta}=S^{i\alpha}_{j\beta} - M^{i\alpha}_{j\beta},
\end{eqnarray}
where $S^{i\alpha}_{j\beta}=M^{k\epsilon}_{l\sigma}U_{k\epsilon}E^{l\sigma}(j\beta)\delta^{i\alpha}_{j\beta}$ are the components of a tensor generalizing the standard degree matrix of a classical network to the case of multilayer, directed and weighted interactions~\cite{de2013mathematical}.
Note that $U \in \mathbm{R}^{N \times L}$ is a rank-2 tensor with all components equal to 1, $E \in \mathbm{R}^{N\times L}$ is a rank-2 canonica tensor with all components equal to 0 except for the entry corresponding to $(j,\beta)$ and $\delta \in \mathbm{R}^{N \times L  \times N \times L}$ is the rank-4 Kronecker delta. 

In practical applications, such tensors are usually flattened via matricization~\cite{kolda2009tensor} into tensors of lower rank, named supra-matrices~\cite{gomez2013diffusion,de2013mathematical}: this operation is useful from an algorithmic perspective, since it conventionally maps a tensor in $\mathbm{R}^{N\times L \times N\times L}$ to a rank-2 tensor in $\mathbm{R}^{NL \times NL}$ where diagonal blocks encode single-layer connectivity, while off-diagonal blocks encode cross-layer relationships.

%% file: framework.tex
\section{Proposed Framework}
\label{s:framework}

In this paper we propose a novel framework to generalize existing Graph Neural Networks (GNNs) to the case of multi-layer networks.
In fact, while the GNN layers have been used successfully to tackle various network science problems, extending their application to the multi-layer and multiplex networks is still an open research question.






\begin{figure*}[!ht]
    \centering
    \includegraphics[width=0.8\textwidth]{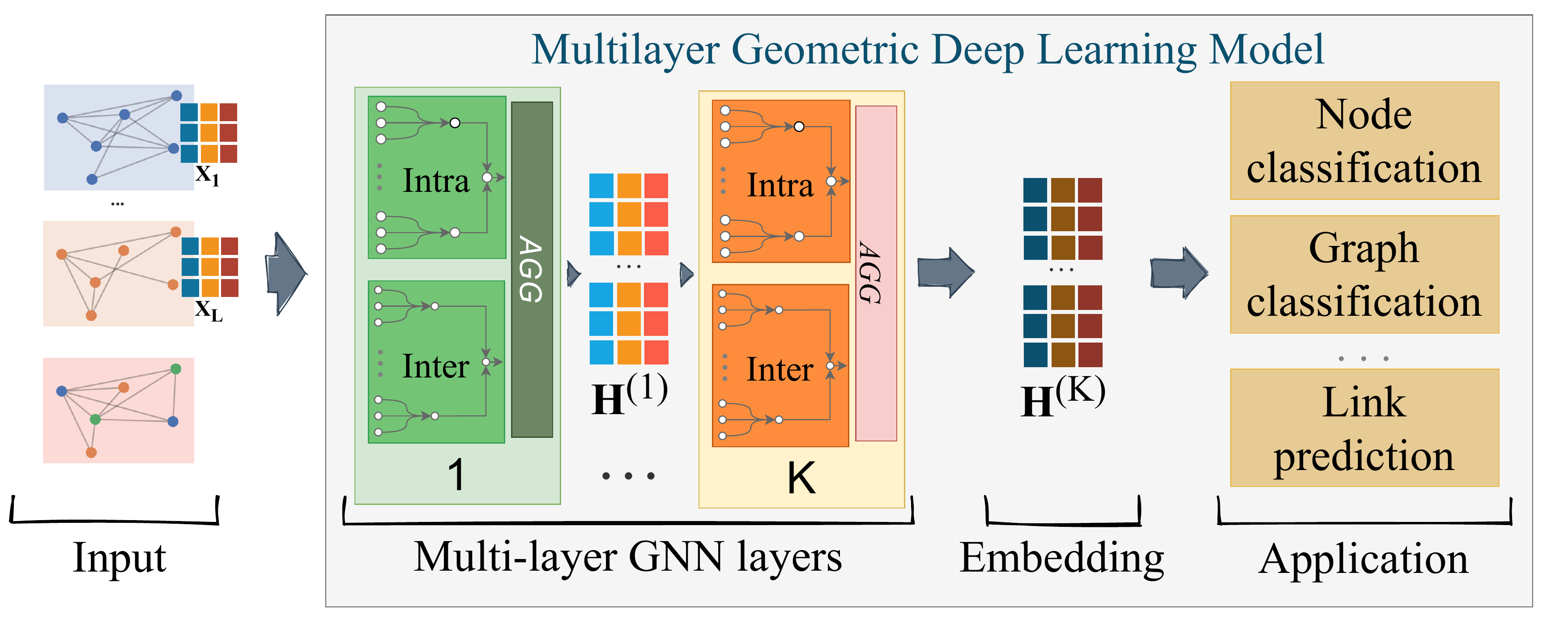}
    \caption{\textbf{A Multilayer Geometric Deep Learning model example.} $K$ is the number of convolutional layers, $\mathbf{X}_{\alpha}$ are the input node features of layer $\alpha$. The node embeddings $\mathbf{H}^{(K)}$, can be used, like in the monoplex case, in various applications.}
    \label{fig:gcn_multilayer}
\end{figure*}

\begin{figure*}[!ht]
    \centering
	    \includegraphics[width=0.50\textwidth]{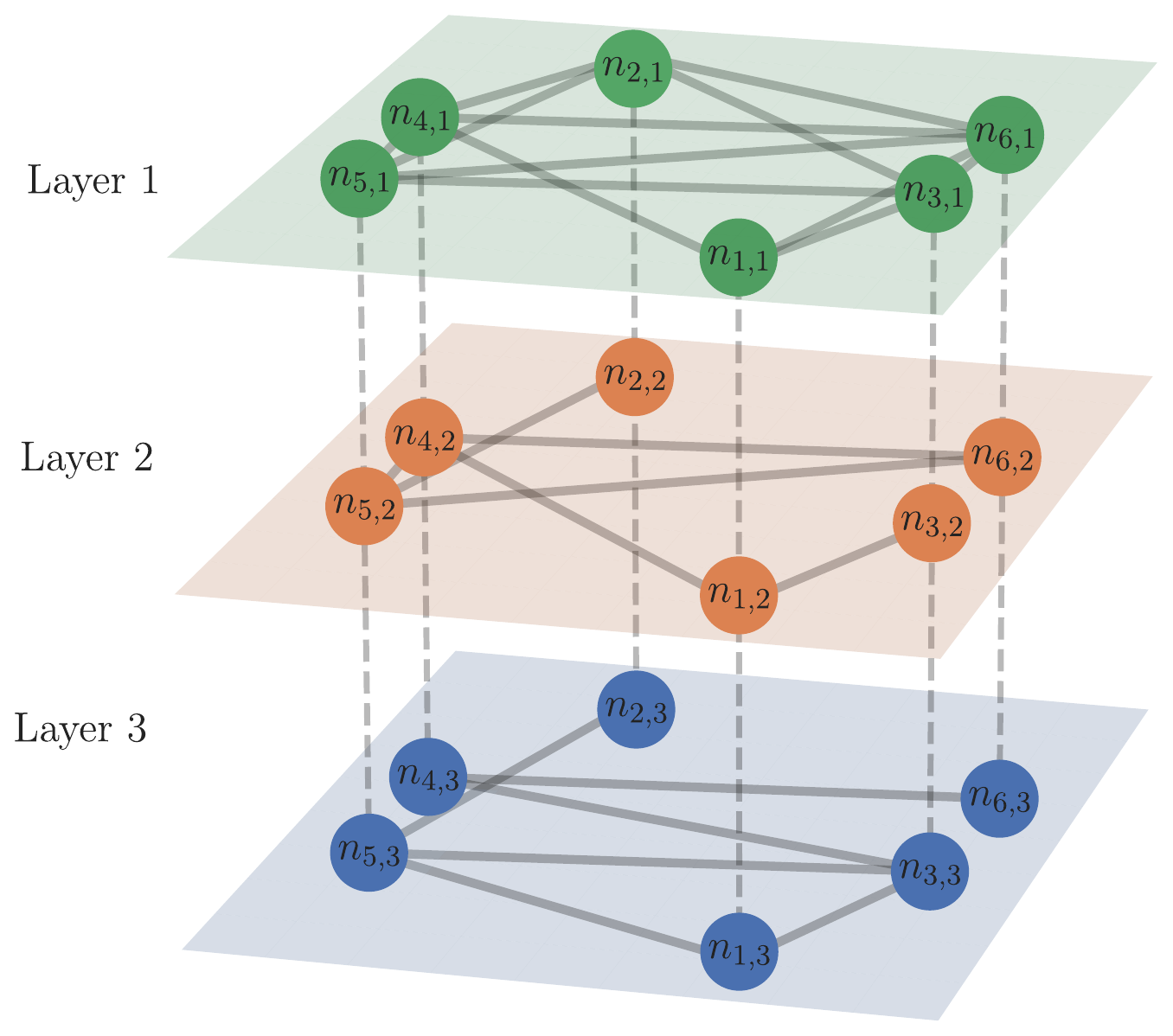}
        \caption{Example multilayer (multiplex) network.}
        \label{fig:multilayer_network}
\end{figure*}

\begin{figure*}[!ht]
    %
    \centering
	\subfloat[\textbf{Step 1:} (virtually) explode the network in \figurename~\ref{fig:multilayer_network} into layers and interlayer links. These sub-networks will be fed to the layers along with the multilayer node features.]{
	    \centering
		\includegraphics[width=0.75\textwidth]{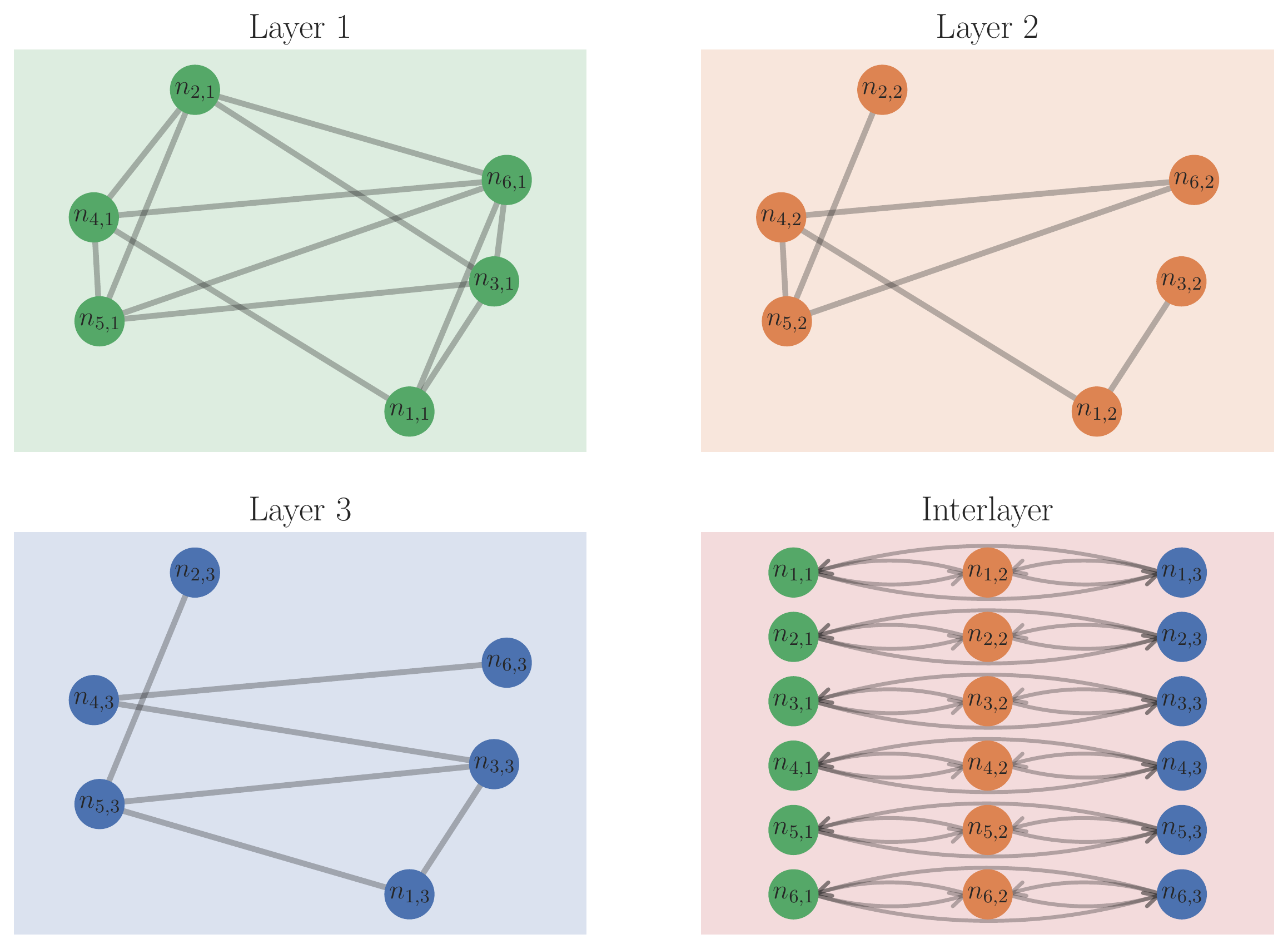}
		\label{fig:multilayer_network_proj}
	}
	\hfill
    \subfloat[\textbf{Step 2:} Computation of multilayer node embeddings. As an example, we show the computation of $\mathbf{h}_{11}^{(k+1)}$ (the embedding of node $1$ of layer $1$) after $k+1$ multiplex-convolutional layers. The intra- and inter- graph convolutional layers work as in the monoplex case (i.e., they propagate the features from the node's neighbourhood), but are provided the multilayer node features as input and the output of the two is aggregated using a generic function, producing the new multilayer node embedding.]{
	    \centering
		\includegraphics[width=0.45\textwidth]{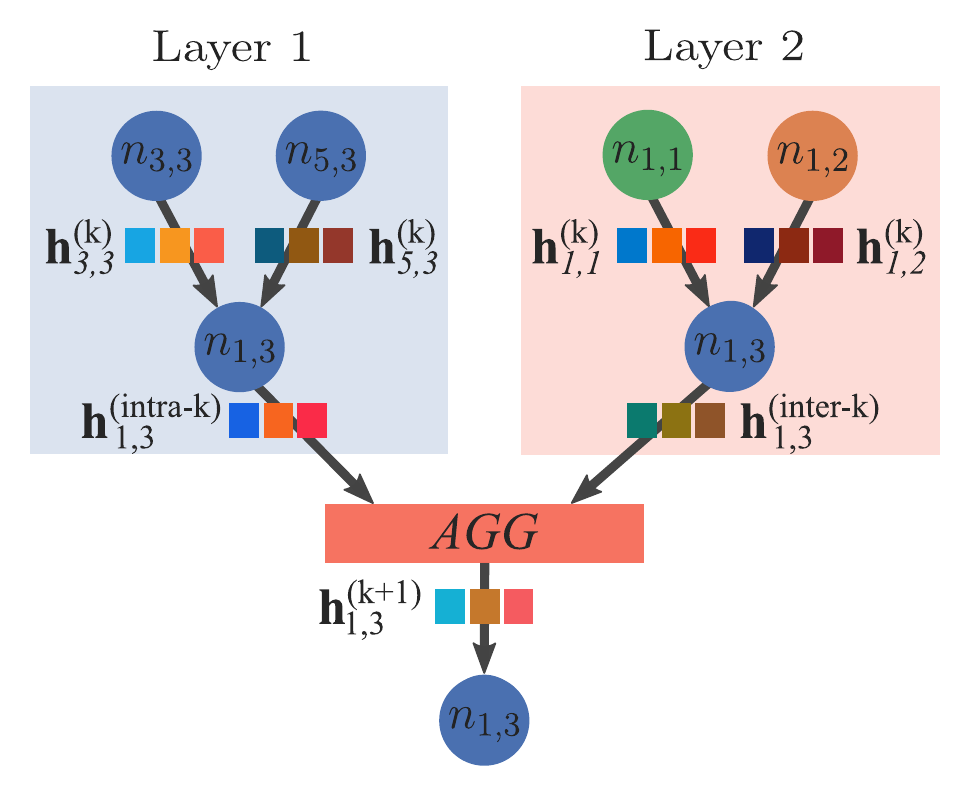}
		\label{fig:embedding_computation}
	}
	\caption{How the multilayer convolutional layers work to produce the multilayer node embeddings.}
    \label{fig:multilayer_example}
\end{figure*}

The framework we propose basically replaces each GNN layer with a \emph{supra-layer} that propagates the node features both in the intra-layer and in the inter-layer neighbourhoods independently --- using at least two different GNN layers --- and then aggregates the two output features to get the embedding of each node.

Merging the notation of the monoplex GNNs, introduced in Section~\ref{s:gnns}, and the notation of the multi-layer networks, introduced in Section~\ref{s:multilayers}, we indicate a node $i$ of network layer $\alpha$ with a tuple $i \alpha$.
In analogy to the mono-plex GNNs, and using such notation, the supra-GNN layer has the following form:

\begin{equation}
    \mathbf{h}_{i \alpha}^{(k+1)} = AGG_{k+1}(\mathbf{h}_{i \alpha}^{(intra-k)}, \mathbf{h}_{i \alpha}^{(inter-k)})
\end{equation}
\begin{equation}
    \mathbf{h}_{i \alpha}^{(0)} = \mathbf{x}_{i \alpha}
\end{equation}

where $\mathbf{h}_{i \alpha}^{(intra-k)}$ and $\mathbf{h}_{i \alpha}^{(inter-k)}$ are computed independently using two different GNN model instances, $AGG_{k+1}$ is some aggregation function (e.g., sum, average, a multi-layer perceptron, etc.) and $\mathbf{x}_{i \alpha}$ are the input features of the node $i$ of layer $\alpha$.

To illustrate the propagation performed by a Graph Neural Network extended to the multilayer case with our framework, we show one of the steps (performed iteratively) of node $n_{1,1}$ of the network in \figurename~\ref{fig:multilayer_network}.
Specifically, a multi-layer network with $L$ layers is first (virtually) exploded into $L + 1$ graphs, where the first $L$ are the layers and the $L+1$-th is the inter-layer network, which connects the nodes of different layers in the multi-layer network and depends on the network type itself.
For instance, if the network is a multiplex, the inter-layer is a network where all the replicas of a node (i.e., the $n_{i\alpha} \ \forall \alpha$) are connected in a clique (i.e., the sub-graph is fully-connected), or, if the network is a multi-layer, the inter-layer connections are the natural connections among the nodes.
After exploding the network and building the (arbitrary) inter-layer graph, node $n_{1,1}$'s embedding after $k+1$ multi-convolutional layers, $\mathbf{h}_{1,1}^{(k+1)}$, are computed as the aggregation (performed by a generic and, possibly, learned $AGG$ function), of its intra- and inter-layer embeddings, $\mathbf{h}_{1,1}^{(intra-k)}$ and $\mathbf{h}_{1,1}^{(inter-k)}$ respectively, which are, in turn, computed using two different GNN layers that work independently but that use $\mathbf{h}_{1,1}^{(k)}$, the multi-layer features from the layer $k$.

The node embeddings $\mathbf{H}^{(k)}$ produced by the supra-layers of the proposed framework (i.e., the features of nodes/replicas) can be then used as common in the monoplex case.
For instance, they can be fed to a Multi-Layer Perceptron (MLP) for regression or classification, or to a pooling layer to perform layer or graph classification.
As an example, if one is interested in node classification in a multiplex setting, the embedding of the replicas can be aggregated using some generic function $ReplicaAGG$ as follows:

\begin{equation}
    \mathbf{\mathbf{r}_n} = ReplicaAGG(\{\mathbf{h}_{n\alpha}^{(k)}, \ \forall \ \alpha \ \in \ L\})
    \label{e:replica_aggregation}
\end{equation}

$\mathbf{\mathbf{r}_n}$ is now the node embedding (i.e., it accounts for all the replicas of node $n$) and can be used for classifying the node.

Such framework has the advantage of decoupling the inter-layer and intra- layer propagation by learning two sets of GNN parameters, enabling the model to learn the different importance of the two propagation "directions", but also allowing the use of different types of GNN layers, which is useful, for instance, to provide intra-layer weights while keeping the inter-layer unweighted (or with default weights).
This overcomes the major limitation of an aggregate-all-layers approach and lets the aggregation weights emerge from the training data.
Moreover, there is no computational complexity overhead as it is the one of the GNN employed (i.e., $O(L \cdot GNN_{intra} + GNN_{inter})$).
In comparison with other multiplex approaches, this framework:
\begin{enumerate*}
    \item is not problem-specific and can work with any multi-layer network (not just multiplexes);
    \item supports any type of training (supervised, unsupervised, reinforcement);
    \item does not need to train the GNNs to reconstruct the network or to define custom loss functions;
    \item supports input node features and propagates them both in the intra- and in the inter- layer, independently;
    \item it allows to stack multiple GNN layers to capture the information in a $K$-hops neighbourhood.
\end{enumerate*}


We stress that the main advantage of this framework is that already existing graph convolutional networks can be used and extended to the multi-layer case with very little effort.
In addition, while we use two identical convolutional layers for each supra-layer, the configuration of the layers is arbitrary, allowing the user to choose different layer parameters or types depending on the specific domain knowledge or application (e.g., the intra-layer is weighted but the inter-layer is not).
This also means that more than a single intra-GNN layer can be used. That is, each supra-layer could include up to one intra-layer GNNs per network layer, if needed to learn different parameters and capture different information or dynamics.

%% file: results.tex
\section{Results}\label{s:results}

To show the validity and generality of the framework proposed in this work, we test it into three different tasks:

\begin{itemize}
    \item Node classification: classification of \emph{var} genes of the human malaria parasite \emph{Plasmodium falciparum};
    \item Link prediction: prediction of (intra-) layer links given a multi-platform social multiplex network, where nodes correspond to users and layers to different social networks;
    \item Network classification: prediction of super-diffusion in multiplex networks.
\end{itemize}

We implement our framework on top of the \emph{PyTorch Geometric}~\cite{Fey/Lenssen/2019} library and manipulate the networks using \emph{graph-tool}~\cite{peixoto-graph-tool-2014}.

\subsection{Malaria genes classification}
In this section, we show how our framework can be used to perform node classification on a biological multiplex network.
Specifically, we use the networks from the work by Larremore et al.~\cite{10.1371/journal.pcbi.1003268}, where they analyze $307$ amino acid sequences from the DBL$\alpha$ domain of the \emph{var} genes of seven \emph{Plasmodium falciparum} isolates.
Two nodes (genes) are connected if they exhibit a pattern of recombination, and authors find nine Highly Variable Regions (HVRs), producing an unweighted and undirected network for each.
Considering that these HVRs share the same set of nodes, we build a multiplex network by using them as layers.
Authors also provide classification of the sequences (nodes) into six classes, based on the number of cysteine residues present in HVR-6, and we use this information to test our framework in a classification task.
Taking into account that in a multiplex network a node appears in all the layers, we reflect this relation in the inter-layer network by connecting all the replicas of each node in a clique.
Regarding the model employed, each of the supra-layers includes a \emph{Graph Attention Network}~\cite{gatlayers} layer that processes all the intra-layers and another one for the inter-layer. The output embeddings of the replicas are transformed into the node embedding as shown in Equation~\ref{e:replica_aggregation}, where $ReplicaAGG$ is a Multi-layer Perceptron.
We train in a supervised manner and test on $20\%$ of nodes, selected using stratified sampling.
That is, the test set contains the same distribution of classes as the full dataset.
Considering that the dataset comes without any node feature, we do not use any as input and assign $\mathbf{x}_{i} = 1$ for replica node $i$.
The final classification accuracy is $83.9\%$, which is remarkable considering the small size of the network, the fact that the classes are strongly unbalanced and that the model is able to classify the nodes using the topology alone without input features.

\subsection{Link prediction}
In this section, we demonstrate how models built according to our framework can be used in the multi-layer link prediction setting.

For this purpose, we build a multi-layer model with \emph{Graph Attention Network}~\cite{gatlayers} layers (one for all the intra-layers and one for the inter-layer, which output are aggregated with a linear layer).
However, compared to the previous example, we replace the node classificator with a link predictor, i.e., a Multi-Layer Perceptron that --- given the embedding of two nodes --- predicts the probability that the two nodes are connected.
Without loss of generality, we perform intra-layer link prediction.
Of course, inter-layer link prediction is still possible.

We test this model on a multiplex social network where nodes, representing users, can interact on three different social networks (FriendFeed, Twitter and YouTube)~\cite{MagnaniSBP10,DBLP:conf/asonam/MagnaniR11}.
More details about the \emph{FF-TW-YT} network layers can be found in Table~\ref{t:ff_tw_yt_network}.
In particular, we train the model to predict the existence of links in one of the layers given the full multiplex network.
We perform the training phase by randomly removing $20\%$ of links from the test layer and by feeding the remaining ones in that layer as positive examples.
The removed links will be used during the test phase to evaluate the performance of the model.
The negative examples are picked randomly in the same quantity among the non-existing ones, both for the train and test phases.
We train and test the models without any input node feature (i.e., $x_i = 1$) on the Twitter and FriendFeed layers.
The test accuracy is $83.3\%$ on the Twitter layer (AUC score $0.829$, F1 score $0.833$) and $81.9\%$ on the FriendFeed layer (AUC score $0.811$, F1 score $0.819$), proving that the model is effectively learning to predict the existence of links.

\begin{table}
\centering
\begin{tabular}{|l|l|l|l|}
\hline
Layer      &    $|N|$       &   $|E|$       & Type          \\
\hline
FriendFeed &    6.4K        &   32.0K       & Directed      \\
Twitter    &    6.4K        &   42.3K       & Directed      \\
YouTube    &    6.4K        &   0.6K        & Undirected    \\
\hline
\end{tabular}
\vspace{0.2cm}
\caption{\emph{FF-TW-YT} network layers.}
\label{t:ff_tw_yt_network}
\end{table}

\subsection{Superdiffusion prediction}

Super-diffusion is a property of certain multiplex networks where the diffusion process is faster than the diffusion on the separate layers~\cite{PhysRevLett.110.028701}, which happens if the second eigenvalue of the supra-Laplacian is greater than the maximum of the ones of the layers.

In this section we demonstrate how our framework can be used to reproduce the results from V.M. Leli et al.~\cite{10.1088/2632-072X/abe6e9}, who predict whether a multiplex network exhibits super-diffusion or not, and do so with a classical Deep Learning model employing CNNs (convolutional neural networks) on the supra-adjacency matrix.
The main limitations of their work are that the models, once trained, are not able to generalize to networks with different number of nodes and layers, and that the models require a lot of training examples.

Predicting whether a multiplex network exhibits the super-diffusion property can be formulated as a network classification task.
For this purpose, use the \emph{Graph Attention Network}~\cite{gatlayers} layers (again, one for all the intra-layers and one for the inter-layer, aggregated with a linear layer) plus a \emph{Global Soft Attention} pooling layer~\cite{li2017gated} that first transforms the node embeddings into layer embeddings (via a learned linear transformation) and then sums them to compute the network prediction.

As in the work from Leli et al., we generate the networks with two layers and $50$ nodes per layer.
The layers are Erd\H{o}s-R\'{e}nyi (ER) networks with $p_i$ for each layer $i$ such that $$0 < p_1 \leq p_2 < 1$$.
We increment the $p_1$ and $p_2$ values in $0.01$ steps and for each combination we generate $5$ networks for the train set and $10$ for the test set.
As a result, we train on $\sim 5K$ networks (we balance the dataset taking all the positive examples and randomly selecting the same number of negative ones), and test on $\sim 50K$ networks.
The final test accuracy is $89.2\%$ (AUC score $0.911$), with the main advantages of exploiting the graph structure and the need of way less networks required to learn and generalize.

%% file: conclusions.tex
\section{Conclusions}
\label{sec:conclusions} 
In this paper we present an innovative way of employing existing Graph Convolutional Networks on multi-layer networks.
Compared to other works, our proposal is problem agnostic and works for any multi-layer network.
Moreover, it is transparent to the training (i.e., any type of training is supported), the node feature propagation happens in both the intra- and inter- layer independently and multiple layers can be stacked to capture information from the topology and the features farther in the network.

We validate our proposal on three different tasks: multiplex node-classification, intra-layer link prediction and network classification.
The results show that the approach is general and performs well in different settings, without any computational over-head which allows the application of the method to large multi-layer networks.